\documentclass[preprint,12pt]{elsarticle}

\usepackage{amssymb}
\usepackage{amsmath,graphicx}
\usepackage{xcolor}
\usepackage{mathtools}
\usepackage{etoolbox}
\usepackage{svg}
\usepackage{csquotes}
\newcommand{\etal}{\textit{et al.}}

\usepackage{multirow}
\usepackage{array}
\newcolumntype{P}[1]{>{\centering\arraybackslash}p{#1}}
\journal{Information Fusion}

\begin{document}

\begin{frontmatter}

\title{\textcolor{black}{Memory Based Fusion for Multi-modal Deep Learning}}

\author{Darshana Priyasad\corref{cor1}}
\ead{dp.don@qut.edu.au}

\author{Tharindu Fernando}
\ead{t.warnakulasuriya@qut.edu.au}

\author{Simon Denman}
\ead{s.denman@qut.edu.au}

\author{Sridha Sridharan}
\ead{s.sridharan@qut.edu.au}

\author{Clinton Fookes}
\ead{c.fookes@qut.edu.au}

\cortext[cor1]{Corresponding author}
\address {SAIVT, Queensland University of Technology, Brisbane, Australia}

\begin{abstract}

The use of multi-modal data for deep machine learning has shown promise when compared to uni-modal approaches with fusion of multi-modal features resulting in improved performance in several applications. However, most state-of-the-art methods use naive fusion which processes feature streams independently, ignoring possible long-term dependencies within the data during fusion. In this paper, we present a novel Memory based Attentive Fusion layer, which fuses modes by incorporating both the current features and long-term dependencies in the data, thus allowing the model to understand the relative importance of modes over time. We introduce an explicit memory block within the fusion layer which stores features containing long-term dependencies of the fused data. The feature inputs from uni-modal encoders are fused through attentive composition and transformation followed by naive fusion of the resultant memory derived features with layer inputs. Following state-of-the-art methods, we have evaluated the performance and the generalizability of the proposed fusion approach on two different datasets with different modalities. In our experiments, we replace the naive fusion layer in benchmark networks with our proposed layer to enable a fair comparison. Experimental results indicate that the MBAF layer can generalise across different modalities and networks to enhance fusion and improve performance.

\end{abstract}


\end{frontmatter}

\section{Introduction}

Multi-modal deep learning has become a major research area with an increasing number of applications that generate/use multiple modalities such as autonomous driving \cite{guan2019fusion, cui2019multimodal, cao2019pedestrian}, emotion analysis \cite{hossain2019emotion, qian2019ears, jiang2020snapshot, guo2019affective}, image fusion \cite{ma2019infrared, zhang2020ifcnn, ma2020infrared} and biometrics \cite{uddin2017multiq, singh2019comprehensive}. Researchers have sought to develop different approaches to combine learned features from different modalities and obtain a common feature space that maximizes the overall performance of the system \cite{calhoun2016multimodal, hori2017attention, poria2017review}. Careful selection of the fusion stage, the model, and it's parameters has enabled researchers to obtain higher performance compared to using a single mode. However, difficulties associated with fusion have created new challenges, and thus fusion has emerged as a separate field of research.

The majority of prior multi-modal fusion research has used naive approaches such as feature concatenation and summation, or fusion methods using attention to filter out uninformative features from the combined space \cite{ghassemian2016review, sun2017multi}. However, with a naive fusion approach, the fusion process only uses information from the current time step and ignores historic information which can play a vital role in multi-modal fusion by means of capturing long-term dependencies between the data domains. With multiple modalities, information from one modality may be occluded or corrupted, leading to an information loss that adversely affects the fusion. However, reasonable adjustments can be inferred by understanding the relationships between the modes and the events captured within them. 

Long-Short Term Memory (LSTM) networks provide a mechanism to capture historical relationships for fusion \cite{gammulle2017two, nweke2019data}. Despite their feedback connections and ability to process sequential data, these methods have limited capacity, especially when considering long-term dependencies. Memory Networks \cite{weston2014memory}, however, are capable of learning long-term dependencies, especially in sequential data \cite{chen2018sequential, miller2016key}, by means of an explicit memory representation. The original concept of an explicit memory has been improved \cite{graves2016hybrid, kumar2016ask} and used in applications not limited to sequential data \cite{zhang2019information}. 

\textcolor{black}{We are motivated by the fact that memory networks consider historical information explicitly, and thus reduce the chance of forgetting important historic information. Therefore, we argue that by incorporating an external memory to store historic data relevant to the fusion, fusion in future iterations can exploit this information and can lead to better representation and recognition.} However extending existing memory architectures to enable fusion is not trivial and can add more complexity to the network which can degrade performance if not carefully designed. Furthermore, consideration needs to be given to the information stored in the memory, and how best to extract information from the memory to assist decision making.

The main objective of this research is to evaluate the performance of a memory-based architecture for multi-modal fusion. We have proposed an improved memory network that receives multi-modal data and outputs fused features, which incorporates both current and historical data. We conduct experiments on different modalities and architectures and show the superior performance and generalizability of the proposed method.

\section{Literature Review}

Multi-modal deep learning has been extensively used in problems such as emotion recognition \cite{cho2019deep, lee2019context} and autonomous driving \cite{feng2020deep, chowdhuri2019multinet} over the past decade. Different modalities including video, multi-spectral data, and sensor data have been used in these applications \cite{li2019multi, kartsch2018sensor, ramachandram2017deep}, and typically deep networks are  used to learn high-level semantic features from each modality, which are then combined through fusion. State-of-the-art methods in multi-modal fusion will be discussed in Section \ref{subsec:fusion}, and memory networks will be outlined in Section \ref{subsec:memnet}.

\subsection{Deep Multi-Modal Fusion}
\label{subsec:fusion}

The fusion of multi-modal features can be carried out at different depths in a deep neural network, and methods can be broadly categorized into three stages: early, intermediate and late \cite{turk2014multimodal, piras2017information}. A widely used fusion methodology, particularly for intermediate fusion, is concatenation, where the features from different modes are combined via concatenation (horizontal or stacked) \cite{nguyen2017deep, lumini2017overview}, summation, or multiplication.

Haghighat \etal \: presented a customized intermediate feature fusion approach using discriminant correlation analysis for biometric recognition \cite{haghighat2016discriminant}. Their method incorporates class associations into the correlation analysis of the feature sets and performs fusion by maximizing the pairwise correlations across the two feature sets. The proposed method eliminates inter-class correlations and strengthens the intra-class correlations. Dong \etal \: proposed an improved late fusion approach based on matrix factorization \cite{dong2018late}. In general, late fusion leads to performance degradation as the predictions from different features can contradict each other. Such methods leverage a hard constraint on the matrix rank to preserve the consistency of predictions by different features. Kim \etal \: used locally adaptive fusion networks to fuse tri-modal (original image, disparity, and matching cost) confidence features (using scale and attention maps) for stereo matching to reconstruct the geometric configuration of scene \cite{kim2019laf}. They have used attention networks to encode the importance of features and then stacked them together in an adaptive and dynamic fashion. 

\textcolor{black}{Ma \etal \: proposed a Generative Adversarial Network (GAN) based fusion method to preserve important uni-modal information within the fused features \cite{ma2020infrared, ma2019fusiongan}. This allows the networks to learn more robust high-level features and achieve higher performance. Since most complex image fusion techniques are associated with higher computational costs \cite{ma2019infrared}, several researchers have focused on developing image fusion techniques with low computational complexity. Zhan \etal \: proposed a gradient magnitude-based image fusion technique to speed-up the fusion process \cite{zhan2017fast}. Furthermore, they have introduced a weighted map based fusion method \cite{zhan2019multimodal} and a structure-preserving filter to fuse medical images \cite{li2018structure}. The importance of these approaches is the associated low computational cost and competitive performance gain. However, since these fusion techniques are largely based on image characteristics, they cannot be used across different modalities such as text and audio.}

Even though the above-stated feature fusion frameworks achieve significantly better results compared to uni-modal networks, they can introduce irrelevant and redundant features within the fused feature space, which reduces the potential performance improvement. Therefore, feature refinement techniques have been used after fusion.

Pouyanfar \etal \: proposed a residual attention based fusion feature refinement method, where the encoded features from each modality are fused using a weighted Support Vector Machine to handle imbalanced data \cite{pouyanfar2019residual}. Lv \etal \: introduced a feature refinement unit using a combination of a CNN and an RCNN which can correct the network’s own identification errors based on the acquired knowledge, and adapt the RCNN to compensate for the lack of feature extraction in the CNN \cite{lv2019attention}. Park \etal \: introduced feature refinement blocks from multiple stages of a deep net to achieve more accurate prediction \cite{park2017rdfnet}. Feature fusion blocks learn residual features from each modality and their combinations to fully exploit the complementary characteristics in the data. Wang \etal \: suggested that the fusion of global features from two modalities (image and 3D point cloud) blindly would degrade estimation performance \cite{wang2019densefusion}. They have proposed a pixel-wise dense fusion network that effectively combines the extracted features by performing local per-pixel fusion instead of global fusion. 

However, these feature fusion and refinement methods are application-specific and fail to generalize across applications. This illustrates a recent trend of introducing fusion methods specific to individual tasks, rather than pursuing a general fusion architecture.

Beard \etal \: proposed recursive multi-attention with a shared external memory \cite{NIPS2015_5846}, which is updated over several iterations as an alternative to naive attention \cite{beard2018multi}, where attention on one modality exploits other modalities. They generate a context vector representing attended features from all modalities and a vector representing the previous iterations, which is passed to the next iteration. This method can easily be extended for multi-modal fusion; however, it is limited by the simple vector representation used to store and retrieve historical patterns.

Zadeh \etal \: proposed a dynamic fusion graph for multi-modal fusion by defining n-modal dynamics as a hierarchical process \cite{zadeh2018multimodal}. They have suggested that each fusion combination from the mode setting has a contribution towards the final fusion outcome, and derived a \enquote{fusion set} graph. The graphs' node connections are weighted by efficacies (to measure how strong the connection between to mode sets), which indicates the contribution. Wang \etal \: proposed a multi granularity fusion approach to fuse information from attentive and global features \cite{wang2018multi}. Yang \etal \: proposed a dynamic fusion method that randomly repels the representations from less significant data sources for fusion \cite{yang2018dfternet}. This enables the network to dynamically select informative modes of information and eliminate uninformative modes from fusion, increasing performance. 

The majority of multi-modal fusion research has limited the developed methods for using the combination of input features at a given time step as described above. \textcolor{black}{Zadeh \etal \: \cite{zadeh2018memory} proposed Memory Fusion Networks (MFN) which can explicitly account for view-specific and cross-view interactions in multi-modal deep learning. Furthermore, they claim that the proposed Multi-view Gated Memory has superior representation capabilities compared to LSTM memories when capturing historical data. However, the proposed mechanism has only been used to capture historical relationships in uni-modal data, and naive fusion (concatenation) has been used for the fusion of the learned multi-modal features. On the other hand, their proposed approach shows that the richer representation of historical context can be captured through complex architectures compared to conventional LSTMs.} The notion of historical context has been applied to a limited number of the fusion approaches \cite{gammulle2017two, beard2018multi}, but those methods fail to learn long term dependencies in fusion. Therefore, we propose the use of an explicit memory network to address the above limitations in multi-modal fusion. 

\subsection{Memory Networks}
\label{subsec:memnet}

Memory networks utilize a storage block and inference components (reader, writer, and composer) together, and learn to use these components jointly \cite{weston2014memory}. The memory can be read and written to, with the aim of using it for prediction tasks. Memory networks were introduced to alleviate the issue of learning long-term dependencies in sequential data. Compared to a Long Short Term Memory (LSTM) unit which updates an internal fixed-size memory representation, memory networks consider the entire history explicitly, eliminating the chance to forget, and the size of memory becomes a hyper-parameter to tune. Sukhbaatar \etal \: has extended the above idea by introducing a recurrent attention model over the external memory which is trained end-to-end \cite{NIPS2015_5846}. This approach can be applied to realistic settings since it needs less supervision during training.

Rae \etal \: have shown that naive memory networks scale poorly in time and space as memory grows \cite{rae2016scaling}. They have proposed a memory access scheme which is end-to-end differentiable: Sparse Access Memory (SAM). They have shown that this method retains the representative power of naive methods while training efficiently with large memories. Miller \etal \: proposed a key-value memory network for question answering to overcome the limitations of knowledge bases \cite{miller2016key}. Their approach makes document reading viable by using a different encoding in later stages (output) of the memory read and addressing. 

Kumar \etal \: proposed a dynamic memory network for question answering that processes questions and input sequences using episodic memories \cite{kumar2016ask}. An iterative attention process is triggered by input questions and it allows the network to condition the attention based on the history of previous iterations and the inputs. Fernando \etal \: introduced a Tree Memory Network which modeled inter-sequence (long-term) and intra-sequence (short-term) relationships using memory modules \cite{fernando2018tree}. The memory was implemented as a recursive tree structure, compared to a naive approach that uses a sequence of historical states. This shows that memory networks can be applied to a diverse set of applications not limited to natural language processing. Fernando \etal \: have used individual memories on two modes before fusion \cite{fernando2017going}, to help extract salient features for fusion. However, the joint learning of the two modalities does not allow modes to see the information within their histories, and determine how this can be utilized in decision making. 

Even though significant advances have been made in areas of multi-modal fusion and memory networks, the applicability of memory networks to multi-modal fusion is not well explored. To the best of our knowledge, \cite{fernando2018learning,fernando2019memory} is the only work that investigates how history can be used to augment feature fusion. However, no investigation has been made regarding 1) what information from each modality should be stored; 2) how to combine different modalities using fusion, and 3) how to effectively extract information from the stored memories to augment the decision making. Our proposed system uses a memory-based fusion layer to model the relationship between data sources and uses it as supportive data for naive fusion. Our proposed layer addresses the above-mentioned research questions.

\section{Methodology}
\label{sec:methodology}

\begin{figure}[t]
\centering
\includegraphics[width=11cm]{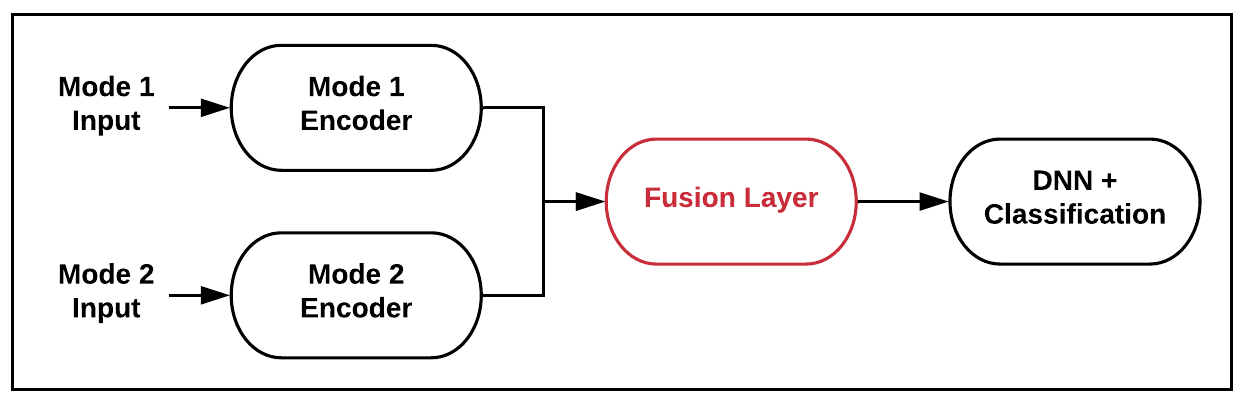}
\caption{\textcolor{black}{High-level system architecture. We have used different encoder networks and DNNs (to capture high-level features), followed by classification for different tasks and modalities. To evaluate the superiority of our proposed fusion architecture, we compare the performance gain achieved by replacing Naive Fusion (NF) with the proposed Memory based Attentive Fusion (MBAF) within the fusion layer (red box), while keeping all other encoder and DNN networks the same throughout a specific task.}}
\label{fig:flowchart}
\end{figure}

\textcolor{black}{In this paper, we propose a novel fusion architecture for multi-modal deep learning with an explicit memory and attention as an alternative to naive fusion. The relative positioning of the proposed layer (red box) in a deep network is illustrated in Figure \ref{fig:flowchart}. We have evaluated the performance gain of the proposed fusion, by changing only the fusion layer while keeping all other components in the overall architecture the same. First, the uni-modal input features are passed through separate feature encoders (see Section \ref{subsec:iemocap} and \ref{subsec:phynet} for the encoder networks used for respective tasks), and then the encoded features are fused by the proposed fusion layer. The resultant features are then passed through a DNN followed by the classification. The main highlight of the paper, the proposed Memory based Attentive Fusion layer (MBAF) is explained in Section \ref{subsec:mem} and the task-specific encoder networks and the DNNs are explained in Section \ref{sec:experiments}.}

\subsection{Mamory based Attentive Fusion (MBAF) Layer}
\label{subsec:mem}

\begin{figure}[tp]
\centering
\includegraphics[width=13.5cm]{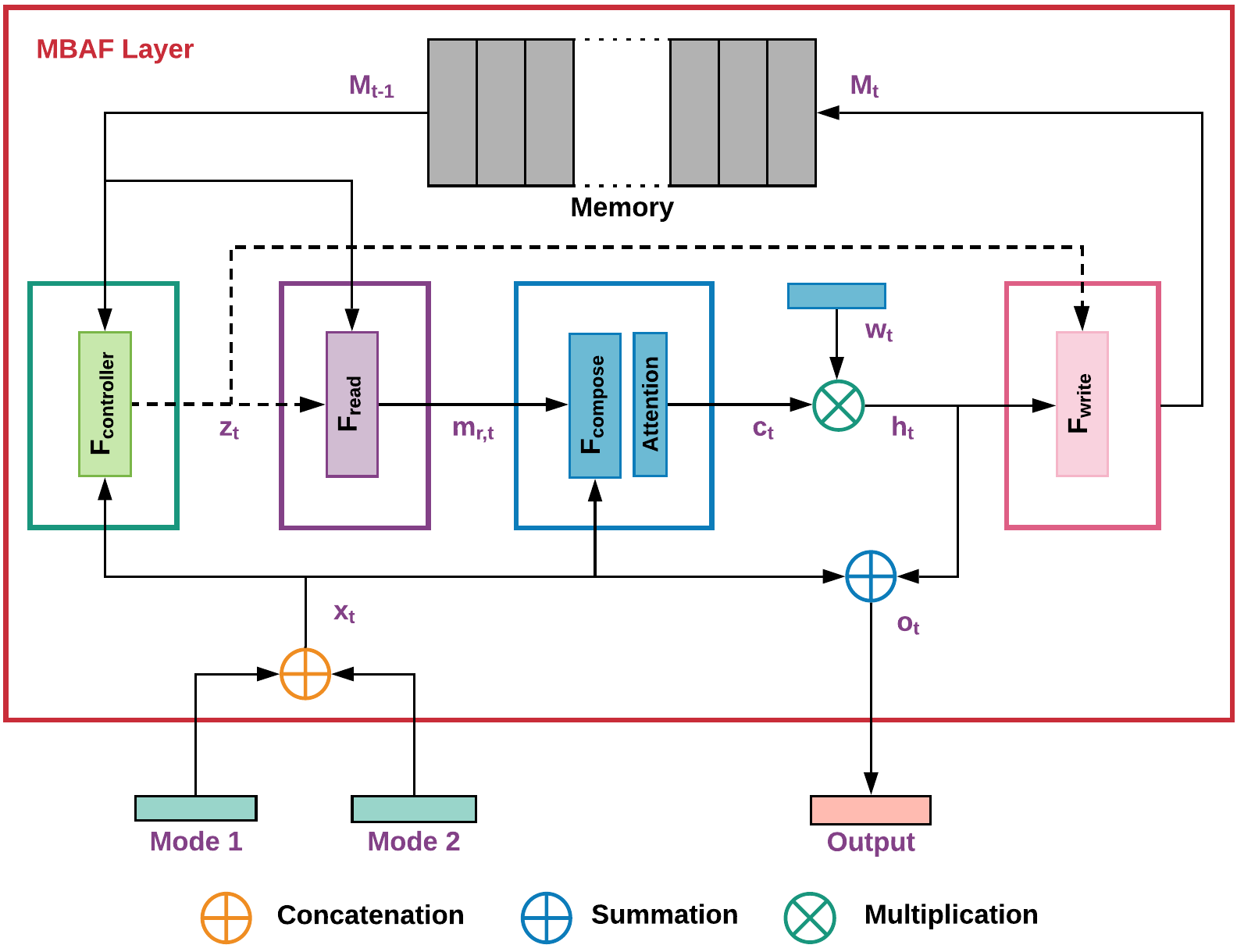}
\caption{Proposed MBAF layer: Inputs are the dense feature vectors from two modalities, and the output is a feature vector of the same dimension as the concatenation of the input features. The inputs are concatenated and the corresponding memory locations' key is calculated (green box). The resultant key is used to read the memory slot \textcolor{black}{(purple box)} and it is fed to the composer (blue box) along with the concatenated input to compose. Self-attention is applied to filter uninformative features from the composed vector. The composer output is transformed (learnable over iterations) and written to the memory slot (pink box) using the pre-calculated key by the controller. The transformed vector is added to the concatenated input and outputs from the layer.}
\label{fig:mbf}
\end{figure}

The proposed Memory-Based Attentive Fusion (MBAF) layer (see Figure \ref{fig:mbf}) consists of four major modules: controller, reader, composer, and writer (the colored boxes in Figure \ref{fig:mbf} refer to these components). The memory is represented by $M \in R^{k,l}$, with a variable number of memory slots where $k$ is the memory length (slots) and $l$ is the hidden dimension. The size of the hidden dimension is set to the shape of the concatenated multi-modal features that are the input to the memory. During initialization, the memory is filled with values from a normal distribution where $\mu=0$ and $\sigma=1$. All other weights are initialized using a uniform distribution. The proposed layer takes dense feature vectors from the two modalities as the input and outputs a tensor with the same shape as naive concatenation to obtain a similar number of parameters for deeper layers and allow fair comparison with naive fusion. 

Consider the dense feature vectors from \textit{mode 1} and \textit{mode 2}, which we denote $x_{m_{1,t}}$ and $x_{m_{2,t}}$ respectively. The memory controller concatenates the feature vectors to obtain $x_{t}$ (Equation \ref{eq:xt}), which is used with the state of the memory from the previous timestep ($M_{t-1}$) to generate a key vector ($z_t$) for a corresponding memory slot which is semantically associated with $x_{t}$. The read module receives $z_t$ and $M_{t-1}$ and retrieves data from the selected slot $m_{r,t}$. The slot location $r$ is defined by $z_t$ which is obtained by attending over memory slots \cite{bahdanau2016end}. The process is defined by Equations \ref{eq:zt} and \ref{eq:mrt}.

\begin{equation}
x_t = x_{m_{1,t}} \oplus x_{m_{2,t}}
\label{eq:xt}
\end{equation}

\begin{equation}
z_t = {\rm softmax} ( f_r\:(x_t)^\top \: M_{t-1})
\label{eq:zt}
\end{equation}

\begin{equation}
m_{r,t} = {z_t}^\top \: M_{t-1}
\label{eq:mrt}
\end{equation}

Similar to the work by Munkhdalai \etal \: \cite{munkhdalai2017neural}, the read function, $f_r$, maps the concatenated dense feature vectors ($x_t$) to the internal memory space and the memory slots related to $x_t$ are determined. The degree of association is calculated and transformed to a key vector $z_t$, and the composing slot, $m_{r,t}$, is determined by the weighted sum of all slots. Then $m_{r,t}$ and $x_t$ are passed to the memory composition component and the composition operations concatenate them followed by passing the representation through a Multilayer Perceptron (MLP), $f^{MLP}_c$, and self-attention. The resultant composition feature vector, $c_t$, is returned. We have used fusion to incorporate both the input and memory in composing the output and an MLP to make the composition learnable. With this, the composition component is capable of learning semantic features from a fused vector space. Attention is applied to the generated features from the MLP to filter uninformative features from the composition. The composition process is defined by Equations \ref{eq:mlp}, \ref{eq:compatt1} and \ref{eq:compatt2}, where $\alpha$ and ${b_t}$ are the attention score and attentive vector respectively. 

\begin{equation}
b_t = f_c^{MLP}( x_t \oplus m_{r,t} )  
\label{eq:mlp}
\end{equation}

\begin{equation}
\alpha_t = {\rm softmax} (b_t)
\label{eq:compatt1}
\end{equation}

\begin{equation}
c_t = \alpha_i\ \otimes b_t
\label{eq:compatt2}
\end{equation}

Then the composition output, $c_t$, is transformed to the encoded memory space, $h_t$, with weight $w_t$ and a ${\rm RelU}$ activation. In the write module, the resulting feature vector, $h_t$, is written to the selected memory slot, $r$, using the pre-calculated key, $z_t$. This updates the memory to the new state, $M_t$. First the data in the corresponding slot is erased and then it is replaced with $h_t$ as indicated in Equations \ref{eq:ot} and \ref{eq:up},

\begin{equation}
h_t = {\rm RelU}\:(c_t \cdot w_t)
\label{eq:ot}
\end{equation}

\begin{equation}
M_t = M_{t-1}(1-(z_t \otimes e_k)^\top) + (h_t \otimes e_t)(z_t \otimes e_k)^\top
\label{eq:up}
\end{equation}

where $1$ is a matrix of ones, $e_l \in R^l$ and $e_k \in R^k$ are vectors of ones and $\otimes$ is the outer product. Finally, the output of the layer ($o_t$) is calculated using $x_t$ and $h_t$ as indicated in Equation \ref{eq:out} where $f^{SUM}_o$ refers to axis wise summation of elements where the output of the MBAF layer is determined by the composition of the current input and previous memory state. This operation is carried out to ensure the output dimension of the layer is similar to that achieved by naive fusion, and the inclusion of memory output $h_t$ ensures that the long term dependencies are incorporated in the fusion operation.

\begin{equation}
o_t = f_o^{SUM} ( x_t \oplus h_t)
\label{eq:out}
\end{equation}

The number of parameters of the proposed fusion layer changes with the input dimensions and it is calculated using,

\begin{equation}
p_{mbaf} = 3(s_{x_{1,t}}+s_{x_{2,t}})^2 + (q+2)(s_{x_{1,t}}+s_{x_{2,t}})
\label{eq:numparam}
\end{equation}

where $p_{mbaf}$, $s_{x_{2,t}}$, $s_{x_{2,t}}$, $q$ refers to the total parameters of the layer, the vector dimension of first and second mode, and the batch size respectively. Similar to the naive fusion layer, this can be extended to $n$ modalities without significant changes to the architecture. Even though this layer has a significantly higher number of parameters (Equation \ref{eq:numparam}) compared to a naive fusion layer, operations are fast and no significant time increase can be observed during inference. 

In the proposed layer, we have developed a novel approach in composition, memory update, and output generation to incorporate historical features and the current input for feature fusion. Compared to other methods, Beard \etal \: \cite{beard2018multi} uses a single vector as the memory (context) while we utilize a memory ($M$) with $l$ slots, which ensures that our model has more capacity to store and retrieve informative facts for different contexts. Fernando \etal \: \cite{fernando2017going} utilized two separate memory layers for two different modalities, ($M^{tr}_t$ and $M^{sp}_t$), while we use a single memory for the combined features. Due to this representation, we are able to build a memory that stores historical relationships among the two modalities that are learned by the network and use that information during fusion, giving the network a better intuition of how modalities behaved in the past compared to the above works. 

In memory composition ($c_t$), both the above works have used naive fusion (concatenation) where we have used an MLP and attention (applied to inputs and memory) to encourage the memory to learn relationships among historical feature representations and current features, and to filter-out uninformative details. In output generation, both the prior methods have used either a transformed $c_t$ (i.e. $h_t$), or $c_t$ itself. However, if the input feature vector contains different information from the historical vectors, it will be augmented and the same information will not be propagated to deeper layers. However, in the proposed method, we use both the input and the transformed features in output generation to minimize the above effect.

\section{Datasets and Experimental Setup}
\label{sec:experiments}

In this section, we describe the datasets used and the experimental setup used to evaluate the performance gain in multi-modal deep learning networks by using the proposed memory enabled fusion, and to establish its generalizability. We have selected benchmarks and baselines from the literature for two different domains: emotion recognition and physiological signal analysis. We have implemented baseline network architectures similar to original works, and replaced the naive fusion layer with the proposed Memory-Based Attentive Fusion (MBAF). The same hyper-parameter configurations were used for both the naive fusion and proposed MBAF to evaluate performance.

\subsection{Experimental Setup for IEMOCAP}
\label{subsec:exp}

The Interactive Emotional Dyadic Motion Capture (IEMOCAP) dataset \footnote[1]{Dataset link : https://sail.usc.edu/iemocap/} for emotion recognition was selected, and our architecture follows the approach of \cite{priyasadattention}. IEMOCAP contains five sessions of utterances for 10 unique speakers along with transcripts. We follow the evaluation protocol of \cite{priyasadattention, yoon2019speech}, and select utterances annotated with four basic emotions: anger, happiness, neutral, and sadness; to achieve an approximately even sample distribution over classes. Samples with excitement are merged with happiness as per \cite{priyasadattention}. The resultant dataset contains $5531$ utterances \{anger:$1103$, happiness:$1636$, neutral:$1708$, sadness:$1084$\}. 

\begin{figure}[t]
\centering
\includegraphics[width=13cm]{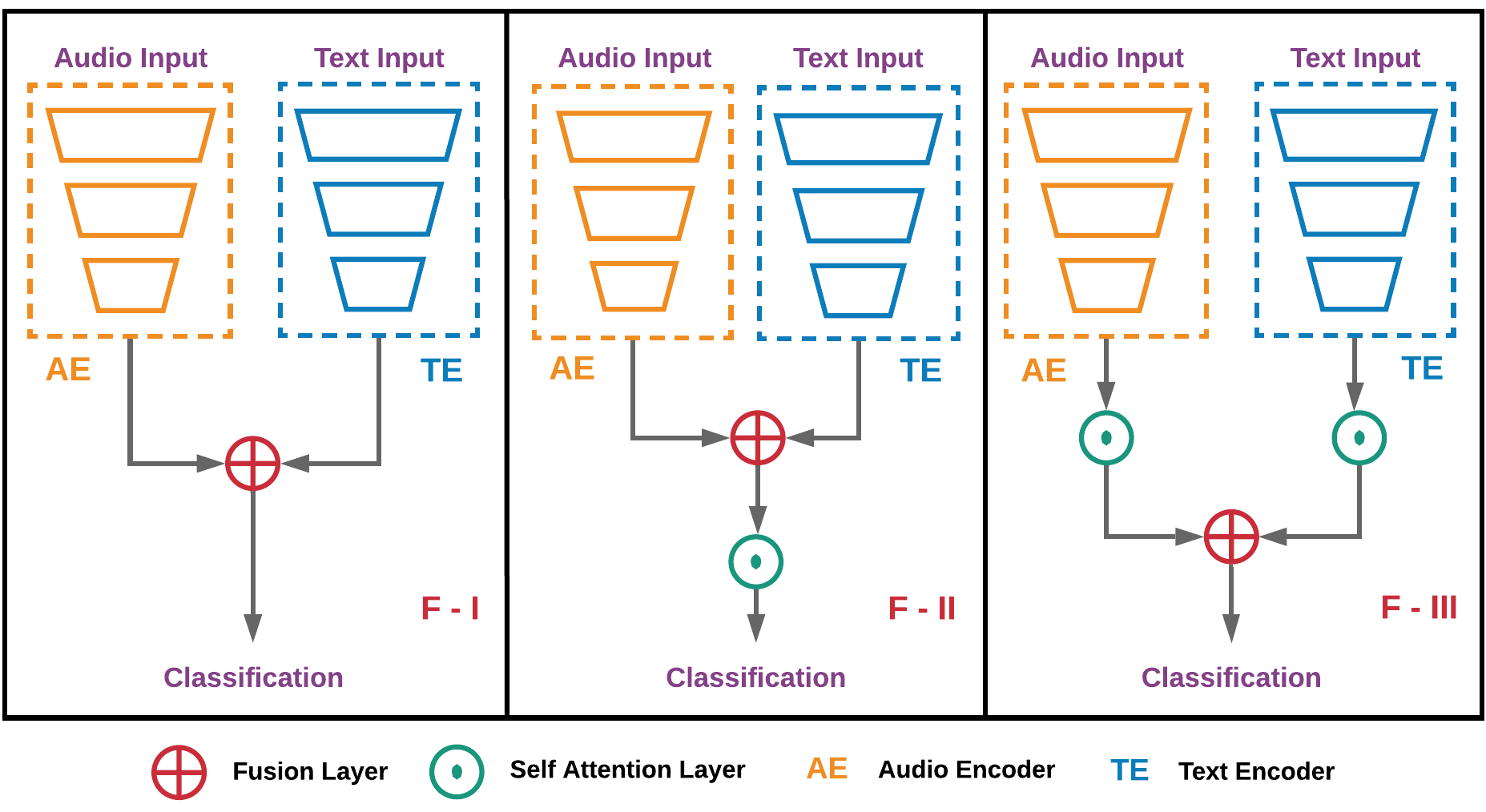}
\caption{Audio and Text inputs are passed through separate encoder networks and the resultant dense features are passed through the memory based attentive network. The encoder networks (AE \& TE) are kept unchanged (see Section \ref{subsec:exp}) from \cite{priyasadattention} where SincNet \cite{ravanelli2018speaker} and a CNN are used for auditory feature extraction, and an RNN and CNN based network is used to extract textual features. Encoded dense features are passed through 3 different fusion networks: \textcolor{black}{F-I, F-II and F-III}. The proposed fusion is applied on text and audio encoded features directly in F-I, while self attention is applied after and before the fusion in F-II and F-III respectively.}
\label{fig:emotionencoder}
\end{figure}

Initial training is carried out on both acoustic (AE) and textual (TE) encoding networks separately before fusion as described below \cite{priyasadattention}. In AE, a $250ms$ audio utterance is passed through a SincNet layer \cite{ravanelli2018speaker} followed by a Deep Convolution Neural Networks (DCNN) with \enquote{Convolution1D} and \enquote{Dense} layers. We obtain a 2048-D feature vector as the output of AE. 

In TE, the input vector is passed through an embedding layer followed by two parallel branches (Left: bi-RNN + DCNN; Right: DCNN) for textual feature extraction. The output vector from the Bi-RNN is passed through three parallel convolutional layers with the filter size of $1$ ,$3$ and $5$; and convolutional layers with the same filter sizes are used in the right branch. Cross-attention is applied to convolution layers with the same filter size from the right branch as the attention for the left branch. The corresponding convolution layers from two branches are concatenated together and passed through a dense layer to obtain a 4800-D feature vector as the output of TE. The resultant dense feature vectors from each modality are then passed through the proposed MBAF layer as illustrated in Figure \ref{fig:emotionencoder}.

Similar configurations have been followed where the sampling rate of each utterance (A), audio segment length (A), window size (A), window shift (A), max sequence length (T), embedding (T) was set to $16000Hz$, $250ms$, $250ms$, $10ms$, $100$ and Glove-300d respectively (A- Audio network configuration, T- Text network configuration). We utilize an 8:1:1 dataset split for training, validation, and testing sets respectively (since the memory access operation depends on the batch size, nearest multiplication of batch size to the split size is selected for test split). The learning rate in each network is fixed at $0.001$ , and the Adam optimizer is used. Experiments were conducted on the baseline architecture with naive fusion and using the proposed fusion. 

\subsection{Experimental Setup for PhysioNet-CMEBS}

We have selected the \enquote{combined measurement of ECG, breathing and seismocardiogram (CMEBS)} dataset \footnote[2]{https://physionet.org/content/cebsdb/1.0.0/} from 
the PhysioNet database. The dataset contains electrocardiogram (ECG), breathing (through respiratory signals - RS) and seismocardiogram (SCG) for 20 healthy people (physiological signals)\cite{garcia2013comparison, PhysioNet}. Subjects are asked to listen to music and the physiological signals were measured at three states; Basel state, while listening to music and after the music. The recordings are 5min, 50min, and 5min in length for each state respectively. The data has been acquired by a Biopac MP36 \cite{garcia2013comparison} where channels 1,2,3 and 4 were devoted to measuring ECG-I, ECG-II, RS, and SCG respectively. Each channel was sampled at 5 kHz.

\begin{figure}[tp]
\centering
\includegraphics[width=10cm]{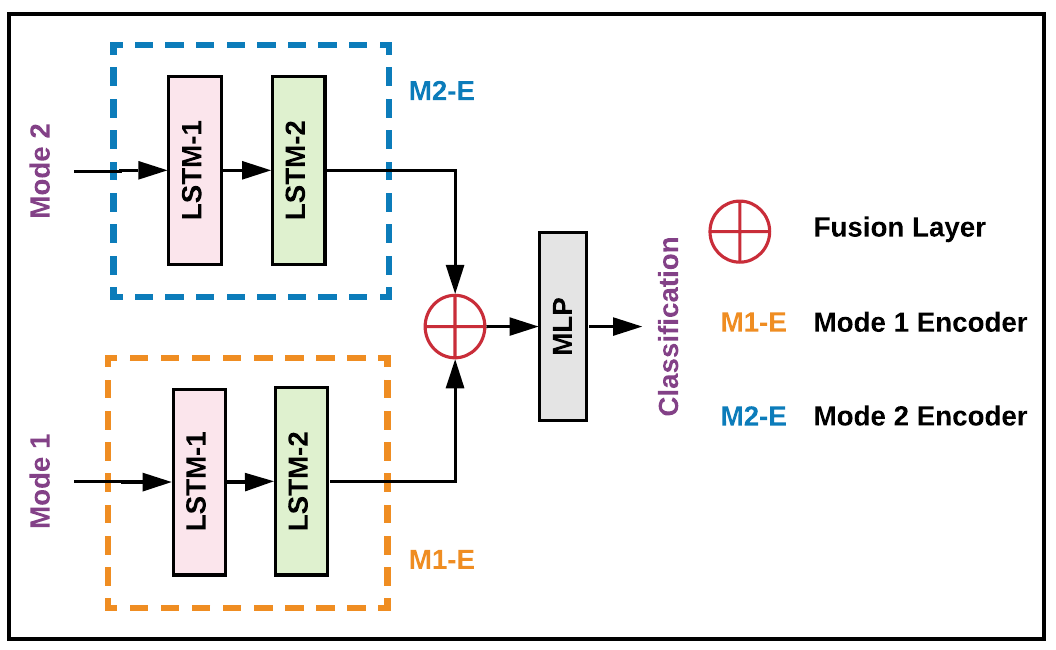}
\caption{Data from two modalities are passed through two LSTMs followed by a fusion layer. The fused features are then passed through a dense layer (MLP) and classified. In evaluations, we use naive fusion and the proposed MBAF in the fusion layer. We use ECG-I, RS and SCG signals in the experiments with combinations of \textcolor{black}{(ECG+RS) as F-IV,(ECG+SCG) as F-V and (RS+SCG) as F-VI.}}
\label{fig:phynet}
\end{figure}

For experimental purposes, we divide and annotate the data according to the acquired state (3 classes). Furthermore, due to the high-class imbalance in the original dataset, we selected a single 5min segment from the second state (first 5 mins). Then each signal was segmented into $200ms$ chunks (each chunk with 1000 data points). The obtained dataset consists of $14,977$ samples \{class 1:\:$4977$, class 2:\:$5000$, class 3:\:$5000$\}. Due to the lack of multi-modal fusion research on this dataset, we created a benchmark model with LSTMs and an MLP as illustrated in Figure \ref{fig:phynet}. We considered 6 consecutive chunks from each modality as a single input to the network. First, the inputs are passed through LSTM-1 with $512$ hidden units which return sequences, followed by LSTM-2 with $512$ hidden units which return a single dense output. The resultant features from both modalities are then fused using the proposed MBAF layer. Then the fused features are passed through a dense layer with $1024$ units and $0.5$ dropout rate followed by a classification dense layer with $3$ units and ${\rm softmax}$ activation. 

Experiments were conducted on the proposed baseline architecture with naive fusion and the proposed fusion. The learning rate and the batch size in each network are fixed at $0.001$ and $32$ respectively, and the Adam optimizer is used. Leave One Subject Group Out (LOSGO) cross-validation is used in this experiment with 4 groups of 5 subjects in each.

\section{Results and Discussion}

The experimental results on both IEMOCAP and PhysioNet-CMEBS datasets are described in Section \ref{subsec:iemocap} and \ref{subsec:phynet} respectively. 

\subsection{Experiments on IEMOCAP Dataset}
\label{subsec:iemocap}

\begin{figure}[tp]
\centering
\includegraphics[width=13.5cm]{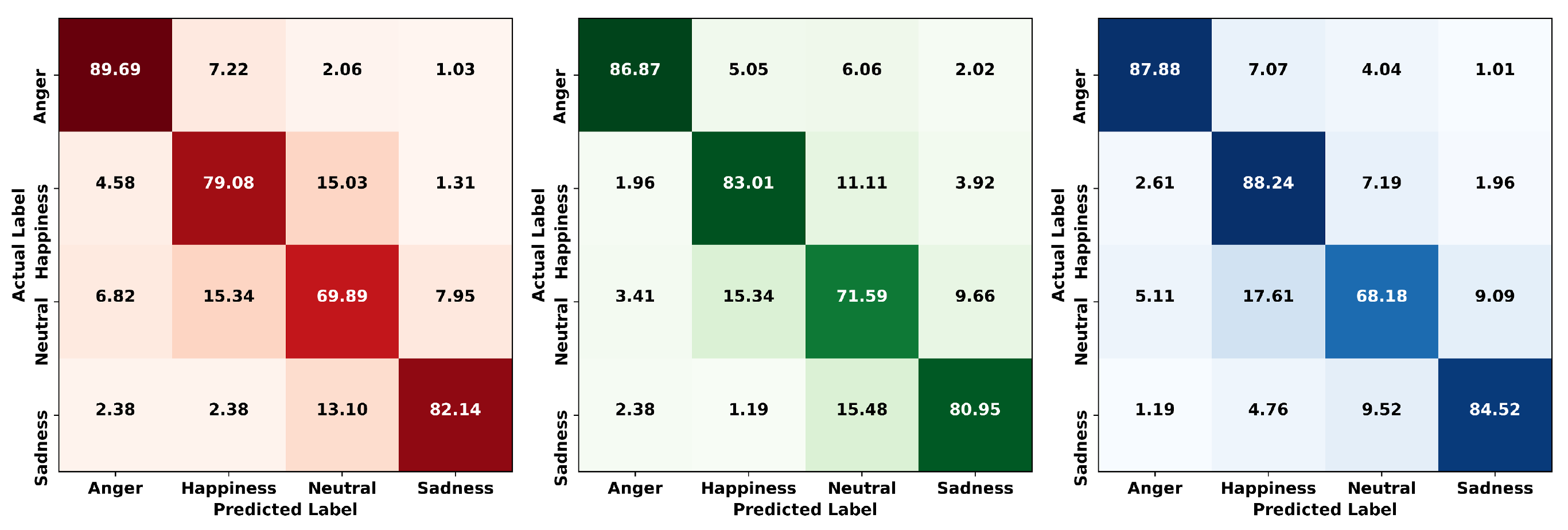}
\caption{Confusion matrices of the proposed fusion layer for separate fusion methods. Left, middle and right figures represent \textcolor{black}{F-I with MBAF, F-II with MBAF and F-III with MBAF respectively.}}
\label{fig:emonetconf}
\end{figure}

Following \cite{priyasadattention, yoon2019speech}, we have measured the performance of our system with weighted accuracy (WA) and unweighted accuracy (UA). Table \ref{tab:iemocapacc} and Figure \ref{fig:emonetconf} presents the performance of our approach for emotion recognition compared with the state-of-the-art methods \footnote[3]{State-of-the-art methods are selected that use the same subset of IEMOCAP dataset (with anger, happiness, neutral and sad emotions) and the same modalities (audio and text only).}. 

\begin{table}[ht]
\caption{\textcolor{black}{Recognition accuracy for IEMOCAP where NF, MBAF, UA and WA refers to Naive Fusion, proposed Memory based Attentive Fusion, unweighted accuracy and weighted accuracy respectively. F-I, F-II and F-III architectures are illustrated in Figure \ref{fig:emotionencoder} and F-III with MBAF has shown the highest recognition accuracy surpassing existing state of the art methods.}}
\vspace{2mm}
\centering
\begin{tabular}{|c|c|c|c|c|} 
\hline 
Model & Modality & WA & UA \\ [0.5ex]
\hline \hline
Conv. MKL \cite{7837868} & $A+T$ & $65.07\%$ &$-$\\
Ens. SVM Trees \cite{rozgic2012ensemble} & $A+T$ & $67.4\%$ & $67.4\%$\\
E-vec+MCNN+LSTM \cite{cho2019deep} & $A+T$ & $64.9\%$ & $65.9\%$\\
MDRE \cite{yoon2018multimodal} & $A+T$ & $71.8\%$ & $-$\\
\textcolor{black}{MHA-1 \cite{yoon2019speech}} & $A+T$ & $75.6\%$ & $76.5\%$\\
MHA-2 \cite{yoon2019speech} & $A+T$ & $76.5\%$ & $77.6\%$\\
\textcolor{black}{F-I (with NF) \cite{priyasadattention}} & $A+T$ & $77.85\%$ & $79.27\%$\\
\textcolor{black}{F-II (with NF) \cite{priyasadattention}} & $A+T$ & $78.98\%$ & $80.01\%$\\
\textcolor{black}{F-III (with NF) \cite{priyasadattention}} & $A+T$ & $79.22\%$ &$80.51\%$\\
\hline
\textcolor{black}{\textbf{Ours F-I (with MBAF)}} & $A+T$ & $78.13\%$ &$79.75\%$\\
\textcolor{black}{\textbf{Ours F-II (with MBAF)}} & $A+T$ & $79.49\%$ &$80.06\%$\\
\textcolor{black}{\textbf{Ours F-III (with MBAF)}} & $A+T$ & $\textbf{80.66\%}$ &$\textbf{82.20\%}$\\
\hline
\end{tabular}
\label{tab:iemocapacc}
\end{table}

\textcolor{black}{Yoon \etal \: has utilized two RNNs to encode textual and acoustic data followed by a DNN for classification in MDRE \cite{yoon2018multimodal}. Cho \etal \: has used two encoders (RNN and a DCNN) to encode two modalities followed by fusion and an SVM for classification \cite{cho2019deep}. MHA-1 and MHA-2 by Yoon \etal \: have used two Bidirectional Recurrent Encoders (BRE) for two modalities followed by multi-hop attention where two different hop counts were used for two methods \cite{yoon2019speech}. Most of the above works have been limited by using hand-crafted features as inputs to uni-modal encoders. However, deep learning has been shown to learn a better feature representation, compared to manually-extracted or calculated features, which has enabled the F-III with NF method from \cite{priyasadattention} to obtain higher results than the above methods. Table \ref{tab:iemocapacc} illustrates that MHA-2 has achieved the highest performance from all the above state-of-the-art methods except for \cite{priyasadattention}. We note that MHA-2 \cite{cho2019deep} and \cite{priyasadattention} are the only methods to use attention for feature refinement, demonstrating the importance of utilizing attention to refine fused features. Priyasad \etal \cite{priyasadattention} has surpassed all the above methods in recognition accuracy via the careful utilization of deep-nets for automated feature extraction and attention. However, all of these methods have used naive fusion (concatenation) as the fusion mechanism ignoring the relative importance of modes over-time.} 

\textcolor{black}{Since F-III with Naive Fusion (NF) by Priyasad \etal \: \cite{priyasadattention} has achieved the best state-of-the-art accuracy, we have selected it as our baseline.} The only architectural differences between our model and the baseline models (F-I, F-II, and F-III with NF) is the proposed memory-based attentive fusion layer, which replaces the naive fusion layer. We have kept all the corresponding layers and parameters the same throughout the experiment. Our proposed model has achieved a substantial improvement in overall accuracy, with a $1.7\%$ increase (in F-III with MBAF) compared to the best baseline F-III with Naive Fusion (NF). Furthermore, we have achieved a performance increase of $0.6\%$ and $0.5\%$ from F-II with MBAF and F-I with MBAF respectively. This performance increase is achieved only by changing the fusion layer. 

\textcolor{black}{It is observed that the network architectures with attention (F-II and F-III) have shown a slight increase in the recognition accuracy compared to non-attentive networks (F-I) with both NF and MBAF fusion. Feature refinement achieved through attention is the main reason behind this observation. Given these results, we argue that the proposed MBAF fusion has allowed the model to understand the relative importance of modes over time by incorporating both the current features and long-term dependencies in the data. Ablation studies have been carried out to determine the impact of different components and hyper-parameters in the explicit memory towards the recognition accuracy, and these studies are presented in Section \ref{sec:ablation}.}

\begin{table}[t]
\caption{\textcolor{black}{Comparison of class-wise Precision, Recall and F1 Score of our proposed approach for the best model (F-III with MBAF) with the state-of-the-art approach of \cite{priyasadattention}. It is observed that the proposed fusion method has achieved a substantial improvement in all the evaluation metrics across the majority of the classes.}}
\vspace{2mm}
\centering
\begin{tabular}{|P{1.8cm}|P{1.5cm}|P{1.5cm}|P{1.2cm}|P{1.5cm}|P{1.5cm}|P{1.2cm}|}
\hline
\multirow{2}{*}{Class}& \multicolumn{3}{c|}{F-III with NF \cite{priyasadattention}} & \multicolumn{3}{c|}{\textbf{F-III with MBAF (Ours)}} \\
\cline{2-7}
& Precision & Recall & F1  & Precision & Recall & F1  \\
\hline
\hline
anger & $87.76\%$ &$86.87\%$ & $87.31\%$ & $\textbf{88.00\%}$ & $\textbf{88.89\%}$ & $\textbf{88.44\%}$\\
happiness & $\textbf{80.00\%}$ &$81.05\%$ & $80.52\%$ & $77.38\%$ & $\textbf{89.97\%}$ & $\textbf{81.00\%}$\\
neutral & $76.19\%$ &$\textbf{72.73\%}$ & $74.42\%$ & $\textbf{81.46\%}$ & $69.89\%$ & $\textbf{75.23\%}$\\
sadness & $74.73\%$ &$80.95\%$ & $77.71\%$ & $\textbf{77.42\%}$ & $\textbf{85.71\%}$ & $\textbf{81.36\%}$\\
\hline
w. avg. & $79.32\%$ &$79.30\%$ & $79.28\%$ & $\textbf{80.84\%}$ & $\textbf{80.66\%}$ & $\textbf{80.51\%}$\\
\hline
\end{tabular}
\label{tab:metric}
\end{table}

\textcolor{black}{Table \ref{tab:metric} presents a class-wise comparison of precision, recall, and F1 score between the two best models: F-III with MBAF (proposed method) and F-III with NF (best method in state-of-the-art). It is observed that the highest precision, recall, and F1 scores have been obtained by F-III with MBAF for all the emotions, except for the highest precision for \enquote{happiness} and highest recall for \enquote{neutral} which are achieved by F-III with NF. Given that the only architectural difference between the two models is the explicit memory, this demonstrates its ability to learn a better decision boundary.}

\textcolor{black}{When comparing our best model, F-III with MBAF with the best model of \cite{priyasadattention} (F-III with NF), substantial improvement in recognition accuracy for \enquote{happiness} and \enquote{neutral} alongside a subtle reduction in \enquote{anger} and \enquote{sadness} is observed. Due to the class imbalance in the dataset, the memory may have focused more on classes with more samples (\enquote{neutral} and \enquote{happiness}; see Section \ref{subsec:iemocap}), resulting in the observed behavior. Since the recognition accuracy of \enquote{anger} remains higher for both the cases, the confusion between other classes which resulted in reduced accuracy is negligible. However, a reduced confusion between \enquote{neutral} with \enquote{happiness} and \enquote{sadness} is observed in F-III with MBAF resulting in an improvement in recognition accuracy of the \enquote{neutral} emotion. The main reason behind this behavior may be that the information captured through the explicit memory is helping to minimize the confusion between \enquote{happiness} and \enquote{sadness} with the \enquote{neutral} emotion.}

\subsection{Experiments on PhysioNet-CMEBS Dataset}
\label{subsec:phynet}

For PhysioNet-CMEBS, we have measured the performance of our system only with weighted accuracy (WA) since the modified dataset has an approximately equal class distribution. Table \ref{tab:physio} presents the performance of our approach for physiological signal state recognition compared with our baseline method. Our proposed model achieves a substantial improvement in overall accuracy, with approximately a $6.5\%$, $2\%$, and $2.5\%$ increase in F-IV (ECG+RS), F-V (ECG+SCG) and F-VI (RS+SCG) respectively, compared to the corresponding baselines using naive fusion.

\begin{table}[tp]
\caption{\textcolor{black}{Recognition accuracy for PhysioNet-CMEBS where NF,MBAF and WA refers to Naive Fusion, proposed Memory based Attentive Fusion and weighted accuracy. F-IV, F-V and F-VI follows the same network architecture with different input modality combinations.}}
\vspace{2mm}
\centering
\begin{tabular}{|c|c|c|c|} 
\hline 
Model & WA-NF & WA-MBAF \\ [0.5ex]
\hline \hline
F-IV (ECG+RS) & $35.23\%$ &$\textbf{41.75\%}$\\
F-V (ECG+SCG) & $51.68\%$ & $\textbf{53.58\%}$\\
F-VI (RS+SCG) & $51.50\%$ &$\textbf{53.98\%}$\\
\hline
\end{tabular}
\label{tab:physio}
\end{table}

Table \ref{tab:runtime} shows the inference time for the above-mentioned models with MBAF compared to the NF for $500$ test samples on a PC with 10 CPU cores and 20GB of memory. Since we have used the same benchmark network with the same number of parameters for F-IV (ECG+RS), F-V (ECG+SCG), and F-VI (RS+SCG) (different modalities with same input dimensions as inputs), we have averaged the inference time for each model. 

A substantial difference in inference time cannot be observed between MBAF and NF for IEMOCAP, even with the high parameter difference. We observe that the model used on the PhysioNet-CMEBS database is much faster, due to the complexity of encoder networks and higher feature dimensions required for processing IEMOCAP. With these results, it is evident that the proposed MBAF layer has a negligible impact on the inference time, even with high dimensional inputs.

\begin{table}[tp]
\caption{\textcolor{black}{Inference time in seconds for the best models with MBAF and NF for 500 samples (time associated with encoder networks are not considered). Even with the significant difference in number of parameters for NF and MBAF for each architecture, MBAF has achieved approximately similar inference times due to associated  simpler calculations.}}
\vspace{2mm}
\centering
\begin{tabular}{|P{2.2cm}|P{2.2cm}|P{1.2cm}|P{1.9cm}|P{2.2cm}|P{1.2cm}|}
\hline
\multirow{2}{*}{Model}& \multicolumn{2}{c|}{with NF} & \multicolumn{3}{c|}{with MBAF} \\
\cline{2-6}
& Parameters & Time & Output & Parameters & Time  \\
\hline
\hline
F-I & $16,420,335$ &$0.49s$ & $6848-D$ &$157,434,371$&$1.74s$\\
F-II & $63,322,307$ &$0.85s$ & $6848-D$ &$204,336,323$&$2.22s$\\
F-III & $43,661,507$ &$0.73s$ & $6848-D$ &$184,675,523$&$1.72s$\\
F-IV - F-VI & $11,448,323$ &$0.96s$ & $1024-D$ &$14,628,867$ &$1.13s$\\
\hline
\end{tabular}
\label{tab:runtime}
\end{table}

\section{Ablation Studies}
\label{sec:ablation}

Several ablations studies have been carried out to identify the; \textcolor{black}{\textbf{1) impact of memory size}, \textbf{2) impact of the location of the memory unit}, \textbf{3) impact of read function on the final classification performance}, \textbf{4) impact of up/down sampling of the memory fusion output towards the classification accuracy and 5) representation of deep-net features}}. The best model (F-III with MBAF) for the IEMOCAP dataset is selected for these experiments and the results are presented in the following subsections. 

\subsection{Impact of Memory Size on Recognition Accuracy}

Table \ref{tab:varmemieomocap} illustrates the variation in recognition accuracy with the number of memory slots, $l$, measured with unweighted and weighted accuracy. Approximately similar weighted accuracies can be observed with lower memory sizes while higher memory sizes achieve comparatively lower accuracies (but higher than naive fusion, refer to Table \ref{tab:iemocapacc}). However, fluctuations are observed with unweighted accuracy where the highest accuracy is obtained with a memory size of $30$.

\begin{table}[ht]
\caption{\textcolor{black}{Variation of recognition accuracy for the best fusion approach (F-III with MBAF) in IEMOCAP with the size of the memory. UA and WA refers to unweighted and weighted accuracy respectively. Maximum recognition accuracy has been observed with a memory size of 30 while the recognition accuracy drops slightly  with higher memory sizes.}}
\vspace{2mm}
\centering
\begin{tabular}{|P{1.8cm}|P{1.4cm}|P{1.4cm}|P{1.4cm}|P{1.4cm}|P{1.4cm}|P{1.4cm}|}
\hline
\multirow{2}{*}{Accuracy}& \multicolumn{6}{c|}{Memory Size} \\
\cline{2-7}
& $10$ & $20$ & $30$ & $40$ & $50$ & $100$ \\
\hline
\hline
WA & $80.07\%$ & $\textbf{80.66\%}$ & $\textbf{80.66\%}$ & $79.10\%$ & $80.07\%$ & $79.10\%$ \\
UA & $81.27\%$ & $81.98\%$ & $\textbf{82.20\%}$ & $80.49\%$ & $81.03\%$ & $80.55\%$ \\
\hline
\end{tabular}
\label{tab:varmemieomocap}
\end{table}

A fluctuating relationship between memory size and the accuracy is observed for the PhysioNet-CMEBS dataset as illustrated in Table \ref{tab:varmemphysio}. 

\begin{table}[ht]
\caption{\textcolor{black}{Variation in weighted recognition accuracy for different fusion methods in PhysioNet as the size of the memory varies. A fluctuating relationship between memory size and the accuracy is observed for the PhysioNet-CMEBS dataset.}}
\vspace{2mm}
\centering
\begin{tabular}{|P{1.8cm}|P{1.4cm}|P{1.4cm}|P{1.4cm}|P{1.4cm}|P{1.4cm}|P{1.4cm}|}
\hline
\multirow{2}{*}{Model}& \multicolumn{6}{c|}{Memory Size} \\
\cline{2-7}
& $10$ & $20$ & $30$ & $40$ & $50$ & $100$ \\
\hline
\hline
F-IV & $39.86\%$ & $\textbf{41.75\%}$ & $41.61\%$ & $40.96\%$ & $40.80\%$ & $41.02\%$ \\
F-V & $\textbf{53.58\%}$ & $53.12\%$ & $52.08\%$ & $52.73\%$ & $52.68\%$ & $52.48\%$\\
F-VI & $52.94\%$ & $\textbf{53.98\%}$ & $52.81\%$ & $52.39\%$ & $51.82\%$ & $53.07\%$ \\
\hline
\end{tabular}
\label{tab:varmemphysio}
\end{table}

Even though the recognition accuracy changes with the memory size, all the recognition rates have surpassed the naive fusion accuracy (refer to Table \ref{tab:iemocapacc} and \ref{tab:physio}). A drop in recognition accuracy can be observed for higher memory sizes since when more history is stored, it is harder for the memory read operation to find salient information. Similarly, when the memory is too small, information is lost and performance degrades slightly. \textcolor{black}{Even though the size of the memory layer increases with the memory slot size, no corresponding increase in inference time was observed. We believe this is due to the simpler tensor operations associated with the selection, retrieval and writing back to memory.}

\subsection{Impact of Read Function on the Recognition Accuracy}

The above described MBAF layer uses an attentive approach to select the memory location in the read module. We have evaluated the performance of using cross-attention over naive attention to retrieve the memory slot since we are inputting two modalities to the layer. For this task, we have defined an additional feature fusion by swapping the order of the fusion ($x_t^s$) and used it to compose ($c_t$) the output which is written back to the memory slot after transformation, as given in Equations \ref{eq:xts}, and \ref{eq:btn}. With this, when the attentive memory selection is carried out, the original fused features ($x_t$) are used as an attention input to the memory which contains transformed features of ($x_t^s$), which behaves as a cross-attention function. All other equations described in Section \ref{sec:methodology} are unchanged.

\begin{equation}
x_t^s = x_{m_{2,t}} \oplus x_{m_{1,t}}
\label{eq:xts}
\end{equation}

\begin{equation}
b_t = f_c^{MLP}( x_t^s \oplus m_{r,t} )  
\label{eq:btn}
\end{equation}

The resulting accuracy variations with the memory size are illustrated in Table \ref{tab:varmemattention}. The results indicate that the use of cross-attention is unable to achieve higher performance compared to general attention. The main reason behind this would be significantly different features in the two modes.

\begin{table}[t]
\caption{\textcolor{black}{Variation of recognition accuracy for the best fusion (F-III with MBAF) in IEMOCAP with size of the memory when the cross-attention (CA) is used to read from memory instead of naive attention (NA). UA and WA refers to unweighted accuracy and weighted accuracy respectively. It is observed that using naive attention in memory composition has resulted in a slight increase of accuracy compared to cross-attention.}}
\vspace{2mm}
\centering
\begin{tabular}{|P{1.8cm}|P{1.4cm}|P{1.4cm}|P{1.4cm}|P{1.4cm}|P{1.4cm}|P{1.4cm}|}
\hline
\multirow{2}{*}{Accuracy}& \multicolumn{6}{c|}{Memory Size} \\
\cline{2-7}
& $10$ & $20$ & $30$ & $40$ & $50$ & $100$ \\
\hline
\hline
WA-NA & $80.07\%$ & $\textbf{80.66\%}$ & $\textbf{80.66\%}$ & $79.10\%$ & $80.07\%$ & $79.10\%$ \\
WA-CA & $79.82\%$ & $\textbf{80.27\%}$ & $79.49\%$ & $78.13\%$ & $79.30\%$ & $79.88\%$ \\
UA-NA & $81.27\%$ & $81.98\%$ & $\textbf{82.20\%}$ & $80.49\%$ & $81.03\%$ & $80.55\%$ \\
UA-CA & $80.80\%$ & $\textbf{81.33\%}$ & $80.45\%$ & $80.22\%$ & $80.99\%$ & $80.93\%$ \\
\hline
\end{tabular}
\label{tab:varmemattention}
\end{table}

\subsection{Impact of Memory Location on Recognition Accuracy}

We have evaluated the impact of the location of the memory-based fusion layer to highlight the importance of using the memory in fusion. For this experiment, we have used the proposed MBAF layer on individual channels by altering Equations \ref{eq:xt} to \ref{eq:xti} as shown. All the other internal calculations were kept the same.

\begin{equation}
x_t = x_{m_{i,t}}\:\: where\:\: i\:\: \in [1, 2]
\label{eq:xti}
\end{equation}

The experiments were carried out for the F-III with MBAF architecture with a memory size of 30 and the accuracy was $78.31\%$ for WA and $80.29\%$ for UA. Using memory for individual modes couldn't outperform the MBAF and NF accuracies. Therefore, it is evident that the long-term dependencies learned through a memory network are capable of increasing the performance of multi-modal fusion over using a naive fusion scheme.

\subsection{Impact of Output Dimension of MBAF layer on Recognition Accuracy}
\label{subsec:out}

\textcolor{black}{The proposed MBAF layer outputs a feature vector ($o_t$) with the same dimension as the naive concatenation output. In F-III with MBAF, two dense feature vectors of 2048-D and 4800-D are input to the layer and a 6848-D feature vector is returned (see Column 6848-D in Table \ref{tab:varoutsize}). However, we can alter the output dimension ($o_t$) of the layer through up or down-sampling. We have evaluated the impact of the dimension of the MBAF layer output on the recognition accuracy and the results are shown in Table \ref{tab:varoutsize}.}  

\textcolor{black}{It is observed that down or up sampling of the original MBAF output (Column 6848-D) has resulted in an accuracy drop. The experiments were carried out for the F-III setting with the MBAF architecture and a memory size of 30. With this result, it is evident that the added up or down sampling has resulted in a reduction of the recognition accuracy.}

\begin{table}[t]
\caption{\textcolor{black}{Variation in recognition accuracy for the best fusion method (F-III with MBAF with memory size $30$) on IEMOCAP as the output dimension of the MBAF layer changes. UA and WA refer to unweighted accuracy and weighted accuracy respectively. The output dimension is changed via up and down sampling and it results in reduction in recognition accuracy.}}
\vspace{2mm}
\centering
\begin{tabular}{|P{2.0cm}|P{1.4cm}|P{1.4cm}|P{1.4cm}|P{1.4cm}|P{1.6cm}|P{1.4cm}|}
\hline
\multirow{2}{*}{Accuracy}& \multirow{2}{*}{6848-D} & \multicolumn{5}{|c|}{Output Dimension of MBAF Layer} \\
\cline{3-7}
& & $512$ & $1024$ & $2048$ & $4096$  & $8192$ \\
\hline
\hline
WA & $\textbf{80.66\%}$ & $76.37\%$ & $76.56\%$ & $75.00\%$ & $76.37\%$ & $73.63\%$ \\
UA & $\textbf{82.20\%}$ & $77.58\%$ & $77.21\%$ & $75.64\%$ & $76.08\%$ & $74.16\%$ \\
Inf. Time & $\textbf{1.72s}$ & $2.02s$ & $2.00s$ & $2.06s$ & $2.21s$ & $2.39s$ \\
\hline
\end{tabular}

\label{tab:varoutsize}
\end{table}

\subsection{\textcolor{black}{Representation of Deep-net Features}}

\begin{figure}[tp]
\centering
\includegraphics[width=13.5cm]{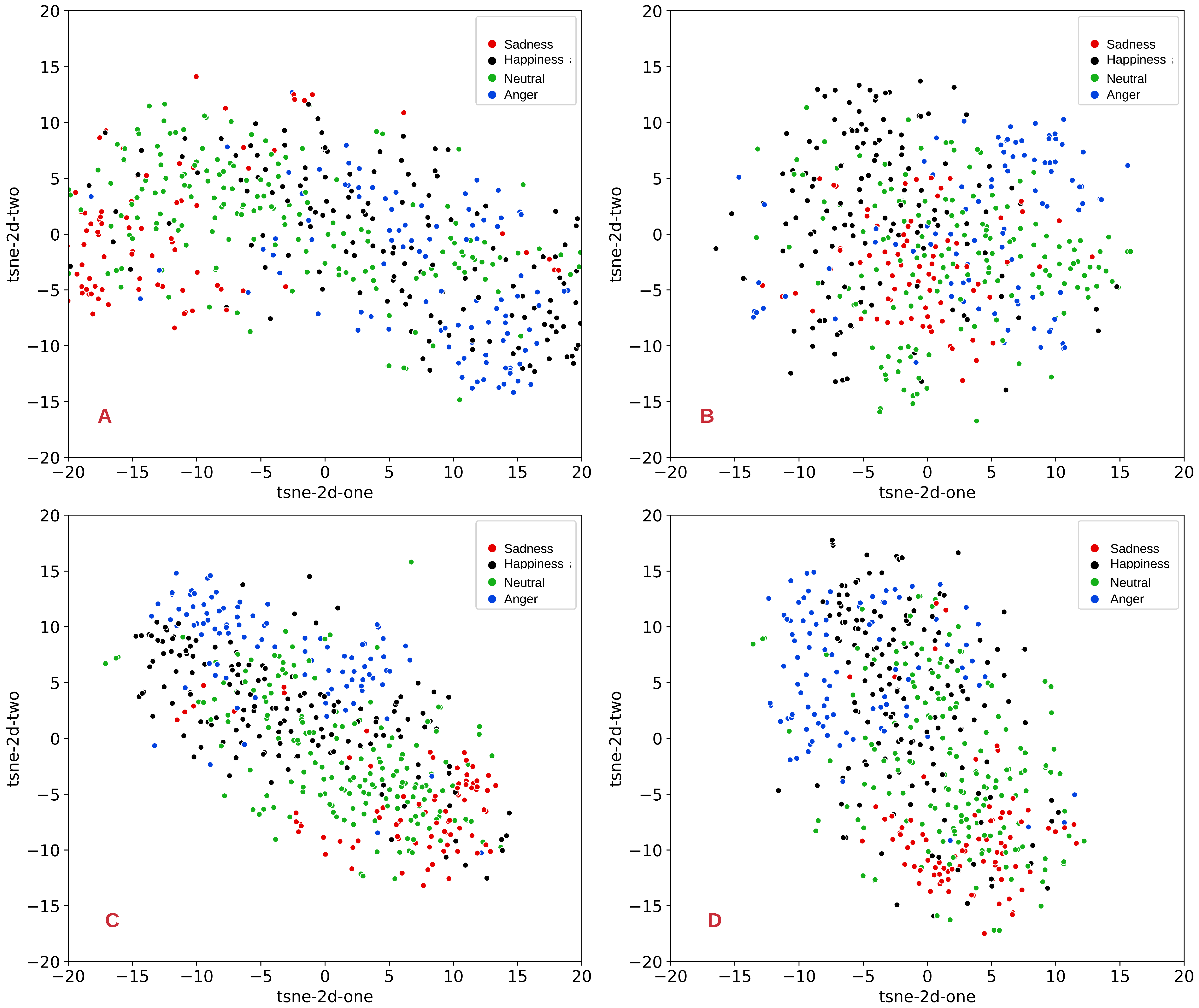}
\caption{\textcolor{black}{t-distributed Stochastic Neighbor Embedding (t-SNE) plot of A) acoustic feature input to fusion layer, B) textual feature input to fusion layer, C) fused features with naive fusion and D) fused features with MBAF for the F-III architecture. The axis limit were manually set to obtain a better visualization ignoring some outlier points observed in B.}}
\label{fig:tnse}
\end{figure}

\textcolor{black}{Figure \ref{fig:tnse} presents a non-linear dimensionality reduction visualization of deep-net features using t-distributed Stochastic Neighbor Embedding (t-SNE) where A, B, C and D refer to acoustic feature input to fusion layer, textual feature input to fusion layer, fused features with naive fusion and fused features with MBAF for F-III architecture respectively. Both uni-modal features were unable to clearly separate any emotion class. Both NF and MBAF fusions have been able to deduce a better decision boundary compared to uni-modal features. Furthermore, we have observed several outlier points in uni-modal features that have been eliminated with fusion suggesting the importance of multi-modal fusion over uni-modal approaches. It is observed that MBAF is capable of learning a better decision boundary for \enquote{sadness} and \enquote{happiness} resulting in reduced confusion with \enquote{neutral} which confirms the results we obtained in Table \ref{fig:emonetconf}. However, we do not observe a perfect decision boundary according to these plots. The main reason for this would be the challenging nature of the problem, and the high dimensionality of the data (6848-D) which has to be reduced to 2-dimensions.}

\section{Conclusion}

\textcolor{black}{In this paper, we propose a novel deep learning architecture for multi-modal data fusion (MBAF layer) based on explicit memory and attention. In contrast to the naive fusion (concatenation) which only considers the features at a given time-step during fusion, MBAF fusion is capable of learning long-term dependencies in data which is combined with the result of naive fusion during the fusion stage. Furthermore, attention over memory composition helps for the refinement of features retrieved from memory. The proposed MBAF layer can be integrated into any deep learning network designed for any task and can be learned end-to-end without having any significant impact on the inference time, even with higher dimensionality in the fusion vector.}

\textcolor{black}{Experiments on publicly available datasets demonstrate the generalizability of the proposed fusion. Furthermore, the experimental results demonstrate that the proposed MBAF is capable of achieving significant improvements, outperforming the naive fusion used by state-of-the-art baselines in terms of classification accuracy. The results indicate that the proposed MBAF fusion allows the model to understand the relative importance of modes over time, by incorporating both the current features and long-term dependencies in the data. Ablation studies show that the proposed fusion methods are capable of achieving stable results over varying hyper-parameters.}

\section{Acknowledgements}

The research presented in this paper was supported partly by an Australian Research Council (ARC) Discovery grant DP170100632.

\bibliographystyle{elsarticle-num}


\end{document}